\documentclass[runningheads]{llncs}

\usepackage{eccv}

\usepackage{eccvabbrv}

\usepackage{graphicx}
\usepackage{booktabs}
\usepackage{bbm}
\usepackage[accsupp]{axessibility}  

\usepackage{hyperref}
\usepackage[table]{xcolor}
\usepackage{multirow}
\newcommand{\cmark}{\textcolor{green!60!black}{$\checkmark$}}
\newcommand{\xmark}{\textcolor{red!75!black}{$\times$}}
\usepackage{orcidlink}

\begin{document}

\title{FineX: Fine-Grained Action Recognition with Cross-Attentive Latent Sparse Experts} 

\titlerunning{FineX: Action Recognition with
Latent Sparse Experts}

\author{Imtiaz Ul Hassan\inst{1}$^{\star}$\orcidlink{0009-0003-3568-3825}
\and
Tasweer Ahmad\inst{1}$^{\star}$\orcidlink{0000-0002-8108-7915} \and
Nik Bessis\inst{1}\orcidlink{0000-0002-6013-3935} \and
Ardhendu Behera\inst{1}$^{\dagger}$\orcidlink{0000-0003-0276-9000}}

\authorrunning{I.~Hassan, et al.}

\institute{Computer Science Department, Edge Hill University, Ormskirk  L39 4QP, United Kingdom 
}

\maketitle
\let\thefootnote\relax\footnotetext{$^{\star}$~Equal contribution.}
\let\thefootnote\relax\footnotetext{$^{\dagger}$~Corresponding author: \href{mailto:beheraa@edgehill.ac.uk}{beheraa@edgehill.ac.uk}}

\begin{abstract}
Fine-grained human action recognition (FHAR) must distinguish visually similar actions that differ mainly in body configuration, timing, or local appearance. RGB representations retain visual context but often suppress joint-level geometry, whereas skeleton representations encode kinematics but discard dense spatial detail. We introduce \textbf{FineX}, which factorizes fine-grained cues into RGB appearance, pose-heatmap geometry, and skeletal-graph topology. Pairwise cross-attention enables symmetric, stream-preserving information exchange, followed by a streamwise latent sparse Mixture-of-Experts that routes each representation to a content-dependent subset of shared experts, regularized by a load-balancing objective. FineX achieves state-of-the-art results on Gym99, Gym288, and Diving48. On the long-tailed Gym288, it raises mean class accuracy from \textbf{68.6\%} to \textbf{76.2\%} (\textbf{+7.6} points) without textual supervision or large-scale vision--language pre-training, demonstrating the benefit of structured visual--pose--graph fusion and conditional expert refinement for FHAR.
\keywords{Fine-Grained Action Recognition \and
Human Behavior Analysis \and
Multimodal Action Recognition \and
Sparse Mixture of Experts}
\end{abstract}

\section{Introduction}
\label{sec:intro}
Fine-grained human action recognition (FHAR) must distinguish action categories that share the same scene, actor, objects, and coarse motion, but differ in subtle body configurations or short temporal phases. Benchmarks like FineGym~\cite{shao2020finegym} and Diving48~\cite{li2018resound} emphasize this: classes often differ only in sub-action order, number of rotations, in-air posture, or slight limb-trajectory variations. Such cues are easily lost with global video descriptors and standard spatio-temporal pooling. FHAR therefore requires representations that retain local body evidence, capture temporal execution, and remain sensitive to fine structural differences among visually similar actions.

Recent work advances FHAR by injecting human pose into video recognition. Pose-guided transformers align joint locations with visual tokens to improve pose–video interaction~\cite{pgvt2024}, while pose-enhanced vision-language models treat pose as an additional modality to adapt large pre-trained representations~\cite{zhang2024pevl}. Recent multimodal methods further exploit RGB-skeleton collaboration and score-level fusion for more discriminative features~\cite{bian2025multimodal}. In parallel, skeleton-based methods focus on fine-grained structural modeling via graph reasoning, decomposition, or hierarchical representation learning~\cite{duan2022revisiting,chang2025hierarchical}, highlighting the importance of body structure for FHAR. However, they still treat “pose” as a single auxiliary representation; joint coordinates, heatmap volumes or skeletal graphs, thus merging distinct fine-grained signals and limiting effective fusion.

We argue that fine-grained action evidence is best understood along three complementary axes. First, RGB appearance captures texture, body orientation, object or equipment contact, and local visual dynamics, but is biased by background context and loses joint-level precision after spatio-temporal aggregation. Second, dense pose heatmaps preserve joint and limb locations in the image plane, providing spatially grounded body configuration while suppressing appearance variation. Third, skeletal graphs encode inter-joint relations and temporal kinematic structure, but discard dense visual and image-plane context. These representations are not interchangeable: heatmaps retain where the body configuration appears, while graphs encode how body parts are topologically and temporally related. Collapsing them into a single pose stream removes a useful distinction for FHAR.

This observation motivates \textbf{FineX}, a cross-attentive, sparse-expert framework for fine-grained action recognition whose representation design explicitly follows the three evidence axes above. We use an R(2+1)D backbone~\cite{tran2018closer} for RGB appearance, a PoseC3D pose-heatmap model with a SlowOnly-R50 backbone~\cite{duan2022revisiting} for dense spatial pose, and ST-GCN++~\cite{duan2022pyskl} for skeletal graph topology. A lightweight fusion module then combines the three streams in two stages. First, pairwise cross-attention lets each stream query the other two while preserving stream identity, so appearance, heatmap, and graph features can exchange evidence without collapsing into a single concatenated representation. Second, a streamwise latent sparse Mixture-of-Experts (MoE) applies conditional nonlinear refinement to each cross-attended stream. Rather than being bound to fixed modalities, a shared expert bank is sparsely selected based on each representation’s content, allowing cue-dependent specialisation to emerge from data.

FineX differs from prior multimodal FHAR methods in both representation and fusion. Rather than treating pose as an auxiliary RGB signal \cite{pgvt2024,zhang2024pevl} or fusing RGB and skeleton features only at the score level~\cite{bian2025multimodal}, FineX separates dense spatial pose from skeletal topology and performs symmetric cross-stream reasoning over all three evidence sources. Unlike dense fusion layers that share one transformation across instances, its sparse expert module refines features per instance, which is crucial because FHAR classes and even samples within a class can depend on different cues: limb angles, body twists, rotation counts, transient poses, or appearance–pose interactions.

\noindent\textbf{Contributions.} 
\textbf{(1)} We formulate FHAR as three-stream evidence decomposition into RGB appearance, dense pose geometry, and skeletal topology.
\textbf{(2)} We introduce symmetric pairwise cross-attention that preserves stream identity while exchanging complementary evidence.
\textbf{(3)} We propose streamwise latent sparse expert routing for content-adaptive refinement without fixed modality--expert assignments.
\textbf{(4)} FineX sets a new state-of-the-art on Gym99, Gym288, and Diving48; on Gym288, it achieves 94.3\% Top-1 and 76.2\% MCA, improving the previous best MCA by 7.6 points.
\vspace{-0.5em}

\section{Related Work} \label{sec:related}
\noindent\textbf{Fine-grained video representation.} Modern action-recognition methods learn spatio-temporal representations via segment aggregation~\cite{tsn2018,tpn2020}, 3D or factorised convolutions~\cite{i3d2017,tran2018closer,slowfast2019,x3d}, and video transformers~\cite{bertasius2021space,arnab2021vivit,fan2021multiscale, li2022mvitv2,videoswin2022,uniformer2022,tong2022videomae}. These models work well when actions differ in scene, object interaction, or coarse motion, but FHAR demands sensitivity to shorter, more local execution patterns. Recent FHAR methods thus add stronger discriminative mechanisms on top of RGB features: DSFNet~\cite{sun2024discriminative} selects informative temporal segments, MANet~\cite{li2025manet} focuses on motion-aware representations, and TAG-Head~\cite{hassan2026tag} augments a video backbone with relational reasoning. While they improve fine-grained discrimination, they still rely mainly on RGB appearance and implicit motion, so detailed body geometry and explicit joint structure must be inferred indirectly from pixels.

\noindent\textbf{Pose and skeleton-based action recognition.} Pose-based methods reduce appearance and background variation by directly modelling body configuration. Graph-based approaches represent joints and bones as a spatio-temporal graph, starting with ST-GCN~\cite{yan2018spatial} and extending to adaptive or stronger variants such as 2s-AGCN~\cite{shi2019two} and ST-GCN++~\cite{duan2022pyskl}. HiOD~\cite{chang2025hierarchical} further improves fine-grained skeleton recognition via hierarchical-aware feature disentanglement. A complementary line uses keypoint heatmaps instead of graph coordinates: PoseC3D~\cite{duan2022revisiting} renders joints and limbs as spatio-temporal heatmap volumes and applies 3D convolutions, gaining robustness to pose noise while preserving image-plane spatial structure. These representations encode different inductive biases: graph models emphasise joint connectivity and kinematic relations, whereas heatmap models preserve dense spatial evidence near joints and limbs. Nevertheless, most pipelines either choose one representation or lump both into a single pose modality, leaving their complementarity under-explored.

\noindent\textbf{Multimodal fine-grained action recognition.} Multimodal FHAR methods increasingly pair RGB video with pose or semantic supervision. PGVT~\cite{pgvt2024} uses pose-guided temporal and spatial attention to link body joints with video tokens. PeVL~\cite{zhang2024pevl} injects pose into a vision--language framework via cross-modal refinement and semantic contrastive loss, extending video--text models such as ActionCLIP~\cite{wang2021actionclip} and X-CLIP~\cite{xclip2022}. MFCF~\cite{bian2025multimodal} strengthens RGB--skeleton collaboration through skeleton-guided localisation, contrastive learning, and score-level fusion. These works show the value of combining appearance and body structure, but fusion is usually restricted to RGB and a single pose-derived stream. They do not jointly model dense pose geometry and explicit skeletal topology as separate, interacting streams, and their pairwise or late fusion lacks a shared mechanism for all streams to selectively query one another.

\noindent\textbf{Adaptive experts and long-tailed video recognition.} Mixture-of-Experts models enable conditional computation by selecting different transformations per input. In long-tailed video recognition, prior work tackles imbalance using class-aware reconstruction and specialised prediction heads~\cite{perrett2023use}, while MEID~\cite{li2023meid} uses multiple experts with internal distillation to improve recognition on imbalanced classes. These approaches highlight the benefit of expert diversity but rely on a single visual representation and do not examine routing over multiple pose and appearance streams. FineX instead couples symmetric pairwise cross-attention with a shared latent expert bank: cross-attention exchanges evidence across RGB, pose-heatmap, and skeletal-graph representations, and sparse routing conditionally refines each stream. This differentiates FineX from both two-stream multimodal fusion and single-representation expert models.
\begin{figure*}[t]
\centering
\includegraphics[width=\textwidth]{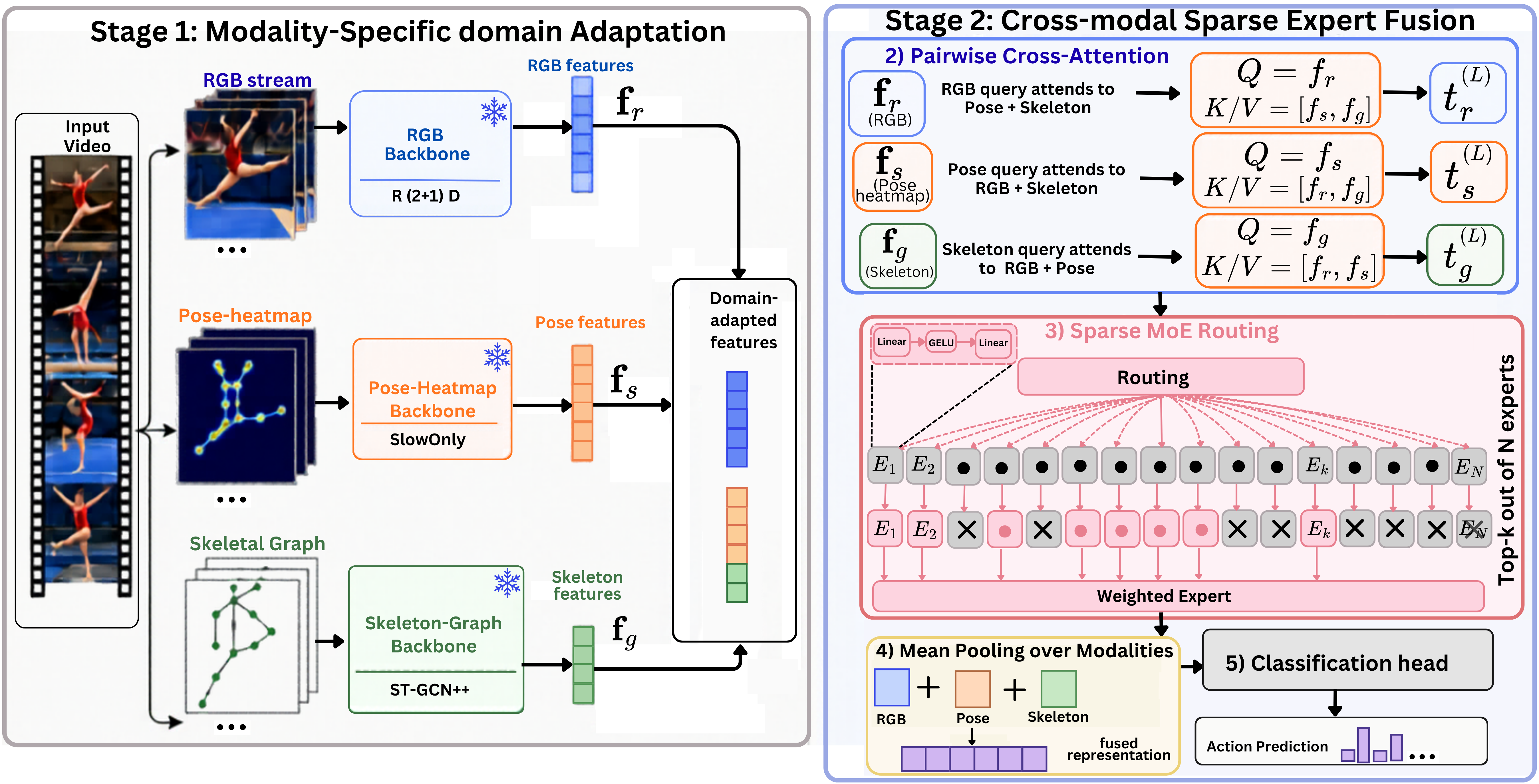}
\caption{\textbf{FineX architecture.} Frozen RGB, pose-heatmap, and skeletal-graph backbones provide pre-cached features, which are projected to a shared space. Pairwise cross-attention contextualises each stream using the other two, after which a shared sparse MoE routes every refined stream representation to its top-$k$ latent experts. The routed outputs are mean-pooled for classification.}
\label{fig:architecture}
\vspace{-2em}
\end{figure*}

\vspace{-0.5em}
\section{Methodology}
\label{sec:method}
Fig.~\ref{fig:architecture} overviews FineX. Three frozen backbones encode RGB appearance, pose-heatmap geometry, and skeletal topology. Their features are projected to a common space, contextualised by pairwise cross-attention, and conditionally refined through a shared sparse MoE before classification.

\noindent \textbf{Problem formulation.}
Consider $S$ video clips $\mathcal{D}=\{(\mathcal{V}_i,y_i)\}_{i=1}^{S}$, where $y_i\in\{1,\ldots,C\}$ is the class label of clip $\mathcal{V}_i$. For each clip, three frozen backbones produce features $\mathbf{f}_m=\mathcal{B}_m(\mathcal{V}_i)$ for $m\in\mathcal{M}=\{r,s,g\}$.
Here, $r, s, g$ denote RGB, pose heatmap, and skeletal graph, with $\mathbf{f}_r,\mathbf{f}_s\in\mathbb{R}^{512}$ and $\mathbf{f}_g\in\mathbb{R}^{256}$. FineX learns
$\mathcal{F}_{\theta}:(\mathbf{f}_r,\mathbf{f}_s,\mathbf{f}_g)\mapsto
\hat{\mathbf{y}}\in\mathbb{R}^{C}$, where $\hat{\mathbf{y}}$ are the predicted class logits. 

\noindent \textbf{Frozen stream representations.} The RGB stream uses R(2+1)D-34~\cite{tran2018closer} initialised from IG-65M~\cite{ghadiyaram2019large}; the pose-heatmap stream uses PoseC3D with a SlowOnly-R50 backbone~\cite{duan2022revisiting}; and the skeletal stream uses ST-GCN++~\cite{duan2022pyskl}. Their penultimate-layer features are pre-cached before fusion training, decoupling the fusion mechanism from further backbone updates and letting the three heterogeneous architectures interface via fixed-dimensional representations.

\subsection{Pairwise Cross-Attention Fusion}
\label{sec:crossattn}
Direct concatenation obscures stream-specific contributions. Instead, we map each feature to a shared dimension $D$:
\(
\mathbf{t}_m^{(0)} = \phi_m(\mathbf{f}_m), \  m \in \mathcal{M},
\)
In layer $\ell \in \{1, \dots, L\}$, each modality $m \in \mathcal{M}$ is contextualised by the other two modalities and is denoted \(\mathbf{C}^{(\ell -1)}_m\).
\begin{equation}
\mathbf{C}_m^{(\ell-1)}
=
\bigl[\mathbf{t}_{m'}^{(\ell-1)}\bigr]_{m'\in\mathcal{M}\setminus\{m\}}
\in\mathbb{R}^{2\times D}.
\label{eq:context}
\end{equation}
Each stream queries this context and is updated using post-normalised multi-head residual attention:
\begin{equation}
\mathbf{t}_m^{(\ell)}
=
\mathrm{LN}\!\left[
\mathbf{t}_m^{(\ell-1)}
+
\mathrm{MHA}^{(\ell)}
\!\left(
\mathbf{t}_m^{(\ell-1)},\mathbf{C}_m^{(\ell-1)},\mathbf{C}_m^{(\ell-1)}
\right)
\right].
\label{eq:update}
\end{equation}
The context is the source for both keys and values, but separate learned projections are used in every head:
\(
\mathbf{Q}_m^{(\ell,h)}=\mathbf{t}_m^{(\ell-1)}\mathbf{W}_Q^{(\ell,h)}\), \(
\mathbf{K}_m^{(\ell,h)}=\mathbf{C}_m^{(\ell-1)}\mathbf{W}_K^{(\ell,h)}\), 
\(\mathbf{V}_m^{(\ell,h)}=\mathbf{C}_m^{(\ell-1)}\mathbf{W}_V^{(\ell,h)}\).
Thus $\mathbf{Q}\in\mathbb{R}^{1\times d_h}$ and
$\mathbf{K},\mathbf{V}\in\mathbb{R}^{2\times d_h}$, where $d_h=D/H$. Head $h$ computes
\begin{equation}
\boldsymbol{\alpha}_m^{(\ell,h)}
=
\mathrm{softmax}\!\left(
\frac{\mathbf{Q}_m^{(\ell,h)}
(\mathbf{K}_m^{(\ell,h)})^\top}{\sqrt{d_h}}
\right),
\qquad
\mathbf{a}_m^{(\ell,h)}
=
\boldsymbol{\alpha}_m^{(\ell,h)}\mathbf{V}_m^{(\ell,h)}.
\label{eq:head_output}
\end{equation}
The $H$ heads are concatenated and projected by $\mathbf{W}_O^{(\ell)}$. Parameters are shared across stream updates within a layer, but each stream supplies a different query; hence the three outputs retrieve different complementary evidence. Stacking $L$ layers yields
$\mathcal{T}^{(L)}=\{\mathbf{t}_r^{(L)},\mathbf{t}_s^{(L)},\mathbf{t}_g^{(L)}\}$.

\subsection{Streamwise Latent Sparse Mixture-of-Experts (MoE)}
\label{sec:moe}
Different fine-grained instances rely on different cues, so a single dense transformation is too restrictive. FineX instead introduces a shared sparse MoE separately to each cross-attended stream. Experts are latent rather than modality-specific: RGB, heatmap, and graph representations share one expert bank, with routing conditioned on each representation’s content.

For sample $b$ and stream $m$, let 
\(
\mathbf{x}_{b,m} = \mathbf{t}_{b,m}^{(L)} \in \mathbb{R}^{D} 
\) denote the final cross attention feature representation to be routed through the sparse MoE. A bias-free linear router predicts logits over $N$ experts:
\begin{equation}
\mathbf{z}_{b,m}=\mathbf{W}_R\mathbf{x}_{b,m}\in\mathbb{R}^{N}.
\label{eq:moe_router}
\end{equation}
The active set $\mathcal{A}_{b,m}=\operatorname{TopK}_{k}(\mathbf{z}_{b,m})$
contains $k$ experts, whose weights are renormalised as
\begin{equation}
\pi_{b,m,i}
=
\frac{\exp(z_{b,m,i})}
{\sum_{j\in\mathcal{A}_{b,m}}\exp(z_{b,m,j})},
\qquad i\in\mathcal{A}_{b,m}.
\label{eq:moe_topk}
\end{equation}
Each expert is a bottleneck feed-forward network,
\begin{equation}
E_i(\mathbf{x})
=
\mathbf{W}_2^{(i)}
\mathrm{Dropout}\!\left(
\mathrm{GELU}(\mathbf{W}_1^{(i)}\mathbf{x})
\right),
\quad
\mathbf{W}_1^{(i)}\!\in\mathbb{R}^{d_e\times D},
\ 
\mathbf{W}_2^{(i)}\!\in\mathbb{R}^{D\times d_e},
\label{eq:expert}
\end{equation}
where $d_e$ is the expert bottleneck dimension. Sparse refinement and stream aggregation are
\begin{equation}
\tilde{\mathbf{t}}_{b,m}
=
\sum_{i\in\mathcal{A}_{b,m}}
\pi_{b,m,i}E_i(\mathbf{x}_{b,m}),
\qquad
\mathbf{u}_b
=
\frac{1}{|\mathcal{M}|}\sum_{m\in\mathcal{M}}\tilde{\mathbf{t}}_{b,m}.
\label{eq:moe_refinement}
\end{equation}
Finally,
$\hat{\mathbf{y}}_b=\mathbf{W}_C\,
\mathrm{Dropout}(\mathrm{LN}(\mathbf{u}_b))+\mathbf{b}_C$.
Cross-attention therefore controls evidence exchange, whereas sparse routing selects the nonlinear refinements applied to each contextualised stream.

\subsection{Training Objective}
\label{sec:loss}
FineX is trained with label-smoothed cross-entropy and an auxiliary load balancing loss for sparse routing, preventing the router from collapsing onto a few experts. For each mini-batch $B$, let $y_b$ and $\hat{\mathbf{y}}_b$ be the ground-truth label and prediction for video $b$, and compute the loss as:
\begin{equation}
\mathcal{L}
=
\frac{1}{B}\sum_{b=1}^{B}
\mathrm{CE}_{\epsilon}(\hat{\mathbf{y}}_b,y_b)
+
\lambda_{\mathrm{lb}}\mathcal{L}_{\mathrm{lb}}.
\label{eq:final_objective}
\end{equation}
Let $\mathcal{R}=\{(b,m):b=1,\ldots,B,\ m\in\mathcal{M}\}$ be the $3B$ routing decisions, $\mathbf{p}_{b,m}=\mathrm{softmax}(\mathbf{z}_{b,m})$, and $\mathcal{A}_{b,m}$ the active top-$k$ set. Following standard sparse-MoE training~\cite{shazeer2017moe}, we define
\begin{equation}
\begin{aligned}
f_i&=\frac{1}{|\mathcal{R}|}\sum_{(b,m)\in\mathcal{R}}
\mathbbm{1}[i\in\mathcal{A}_{b,m}],&
q_i&=\frac{1}{|\mathcal{R}|}\sum_{(b,m)\in\mathcal{R}}p_{b,m,i},\\
\mathcal{L}_{\mathrm{lb}}&=N\sum_{i=1}^{N}f_iq_i .
\end{aligned}
\label{eq:load_balance}
\end{equation}
Here $f_i$ measures expert selection frequency and $q_i$ supplies a differentiable signal through the full router distribution. Their product penalises experts that simultaneously attract excessive assignments and probability mass. We use $\epsilon=0.1$ and $\lambda_{\mathrm{lb}}=0.1$.

\vspace{-0.5em}

\section{Experiments}
\label{sec:experiments}
\subsection{Experimental Setup}
\label{sec:setup}
\noindent\textbf{Datasets and metrics.}
Gym99/Gym288~\cite{shao2020finegym} contain 99/288 gymnastics classes with 20K/23K training and 8.5K/9.6K test clips; Gym288 is highly long-tailed. Diving48~\cite{li2018resound} has 16.1K training and 2.3K test clips from 48 dive classes. We use official splits and report Top-1 on all datasets and MCA on Gym99/Gym288.

\noindent\textbf{Pose and training protocol.}
We use PySKL HRNet keypoints~\cite{sun2019deep,duan2022pyskl} for Gym99, and D-FINE~\cite{peng2025d} plus ViTPose-L~\cite{xu2022vitpose} for Gym288 and Diving48, keeping the largest person per frame. Stage~1 adapts the three backbones independently; Stage~2 freezes them and trains FineX on pre-cached features to isolate fusion gains. Unless noted, $D{=}512$, $L{=}3$, $H{=}8$, $N{=}16$, $d_e{=}256$, and $k{=}8$. The fusion module is trained for up to 80 epochs with Adam (base learning rate $3{\times}10^{-4}$, weight decay $10^{-3}$), cosine scheduling, gradient clipping, label smoothing, and Eq.~\eqref{eq:load_balance}. Further implementation details are in the supplement.

\vspace{-0.8em}
\subsection{Comparison with State-of-the-Art (SOTA)}
\label{sec:sota}
\begin{table*}[t]
\centering
\caption{\textbf{Comparison with SOTA}. Top-1 accuracy and Mean Class Accuracy (MCA) in \% on Gym99/Gym288, and Top-1 accuracy on Diving48. R $\rightarrow$ RGB, T $\rightarrow$ Text, P $\rightarrow$ Pose.} 
\label{tab:merged_comparison}
\scriptsize
\setlength{\tabcolsep}{2pt}
\begin{tabular}{|l|l|l|cc|cc|c|}
\hline
\textbf{Model} & \textbf{Venue} & \textbf{Input} 
& \multicolumn{2}{c|}{\textbf{Gym99}} 
& \multicolumn{2}{c|}{\textbf{Gym288}} 
& \textbf{Diving48} \\
\cline{4-8}
& & 
& Top-1 & MCA 
& Top-1 & MCA 
& Top-1 \\
\hline
 
TSN\cite{tsn2018} & TPAMI’18 & R & 74.8 & 61.4 & 68.3 & 26.5 & -- \\
I3D\cite{i3d2017} & CVPR'17 & R & 75.6 & 64.4 & 66.1 & 28.2 & -- \\
TRNms\cite{trn2018} & ECCV'18 & R & 79.5 & 68.8 & 73.1 & 32.0 & -- \\
ST-GCN++$^\dagger$\cite{duan2022pyskl} & ACM MM'22 & P & 94.2 & 91.9 & 85.9 & 61.3 & 85.9 \\
TSM\cite{tsm2019} & ICCV'19 & R & 80.4 & 70.6 & 73.5 & 34.8 & -- \\
PoseC3D$^\dagger$\cite{duan2022revisiting} & CVPR'22 & P & 94.4 & 92.0 & 87.2 & 61.4 & 82.8 \\
SlowFast\cite{slowfast2019} & ICCV'19 & R & 93.9 & 90.6 & 86.8 & 51.2 & -- \\
TQN\cite{tqn2021} & CVPR'21 & R & 93.8 & 90.6 & 89.6 & 61.9 & 81.8 \\
 
DSFNet\cite{sun2024discriminative} & TOMM'24 & R & 90.3 & 86.3 & 84.9 & 47.6 & 85.0 \\
MANet\cite{li2025manet} & CIS'25 & R & -- & -- & -- & -- & 81.8 \\
TAG-Head\cite{hassan2026tag} & ICPR'26 & R & 95.6 & 93.8 & 92.2 & 68.6 & -- \\

HiOD\cite{chang2025hierarchical} & ICCV'25 &  P & 96.3 & -- & -- & -- & -- \\

\hline


PGVT\cite{pgvt2024} & WACV'24 & R+P & 96.7 & 91.6 & 91.0 & 63.6 & 91.3 \\
MFCF\cite{bian2025multimodal} & BMVC'25 & R+P & 96.0 & -- & -- & -- & -- \\
 
\hline
 
PeVL\cite{zhang2024pevl} & CVPR'24 & R+T+P & 97.0 & 91.8 & 91.8 & 64.0 & 92.5 \\
\hline
 
\textbf{Ours} & -- & \textbf{R+P} 
& \textbf{97.1} & \textbf{96.4} 
& \textbf{94.3} & \textbf{76.2}
& \textbf{92.9} \\
\hline
 
\end{tabular}
\par\smallskip

{\footnotesize $^\dagger$ ST-GCN++ and PoseC3D (SlowOnly backbone) were trained from scratch using the pose data extracted by our pipeline.}
\vspace{-1.5em}
\end{table*}

Table~\ref{tab:merged_comparison} compares FineX with representative RGB, pose-based, multimodal, and vision--language methods. We report Top-1 accuracy for all datasets and MCA for Gym99/Gym288, where class imbalance makes sample-averaged accuracy inadequate. FineX achieves the best performance on all benchmarks, with the largest gains on Gym288, the most fine-grained and long-tailed setting.

\noindent
\textbf{Gym288:} Gym288 is the most diagnostic benchmark in Table~\ref{tab:merged_comparison} because it combines 288 fine-grained classes with a highly long-tailed distribution. FineX achieves \textbf{94.3\%} Top-1 and \textbf{76.2\%} MCA, surpassing the strongest prior MCA result, TAG-Head~\cite{hassan2026tag}, by \textbf{+7.6} points and the vision--language baseline PeVL~\cite{zhang2024pevl} by \textbf{+12.2} points. The MCA gain is crucial: Top-1 is dominated by frequent classes, whereas MCA weights classes equally and is more sensitive to tail performance. This is reflected in the Top-1--MCA gap: TAG-Head and PeVL reach high Top-1 accuracy but still have gaps of 23.6 and 27.8 points, respectively. FineX reduces this gap to \textbf{18.1} points, the smallest among methods reporting both metrics on Gym288, indicating better class-balance rather than improvements limited to head classes.

These results support FineX’s core design. RGB-only methods such as TQN~\cite{tqn2021} and TAG-Head~\cite{hassan2026tag} offer strong appearance and motion cues but lose joint-level spatial detail after spatio-temporal aggregation. Pose-only models such as PoseC3D and ST-GCN++ preserve body structure but lack appearance context to distinguish visually similar actions. Multimodal methods such as PGVT~\cite{pgvt2024} and PeVL~\cite{zhang2024pevl} improve over single-stream baselines, yet do not explicitly disentangle dense pose geometry from skeletal graph topology. FineX fuses these complementary signals with stream-preserving cross-attention and then applies content-adaptive sparse expert refinement, achieving the largest gains where single-stream evidence is weakest. The per-class analysis in Fig.~\ref{fig:longtail_gain} shows that these gains are concentrated on rare and mid-frequency classes.

\noindent
\textbf{Gym99:} On Gym99, FineX reaches \textbf{97.1\%} Top-1 and \textbf{96.4\%} MCA (Table~\ref{tab:merged_comparison}). Its Top-1 gains are modest, \textbf{+0.1} over PeVL~\cite{zhang2024pevl} and \textbf{+0.8} over HiOD~\cite{chang2025hierarchical}, indicating that Gym99 is nearing saturation around 96--97\% Top-1. MCA remains more revealing: FineX outperforms PeVL by \textbf{+4.6} and TAG-Head~\cite{hassan2026tag} by \textbf{+2.6}. Its Top-1--MCA gap is only \textbf{0.7}, smaller than for other methods with both metrics reported, showing that FineX improves recognition consistently across classes rather than only on easy ones. Because Gym99 is less long-tailed than Gym288, these gains indicate that the proposed cross-modal decomposition also enhances fine-grained discrimination in a more balanced setting.
 
\noindent\textbf{Diving48:}
FineX achieves \textbf{92.9\%} Top-1 accuracy, surpassing the vision--language baseline PeVL~\cite{zhang2024pevl} by \textbf{+0.4} and the strongest unimodal pose baseline ST-GCN++ by \textbf{+7.0}. Diving48 reduces scene-context bias: most clips share a similar aquatic setting and differ mainly in take-off mechanics, somersault and twist count, body configuration, and entry posture. In this regime, textual semantics or scene priors add little; the key signal is fine-grained motion and body structure. FineX fits this setting because pose heatmaps preserve image-plane joint geometry, skeletal graphs capture inter-joint kinematics, and RGB features add complementary appearance cues. These results show that structured visual--pose--graph fusion can match or exceed vision--language alignment when recognition hinges on subtle motion execution rather than semantic context.

\noindent Across all datasets, the comparison suggests that FineX improves not by simply increasing model scale, but by aligning the fusion architecture with the evidence required by FHAR. The largest gains appear where fine-grained cues are most difficult to recover from any single stream: long-tailed categories in Gym288 and motion-structured classes in Diving48.

\vspace{-1em} 

\subsection{Ablation Studies}
\label{sec:ablation}

We run ablations on all three benchmarks to quantify the impact of FineX’s main design choices: pairwise cross-attention, streamwise sparse expert routing, cross-stream complementarity, load balancing, and the number of active experts. Unless noted, each experiment varies only one factor while keeping the rest of the training protocol fixed.

\begin{table}[t]
\centering
\caption{\textbf{Component ablation}. Cross-attention exchanges evidence across streams, while sparse MoE performs streamwise conditional refinement. Full FineX uses both.}
\label{tab:ablation_component}
\scriptsize
\setlength{\tabcolsep}{4pt}

\begin{tabular}{|cc|cc|cc|c|}
\hline
\multicolumn{2}{|c|}{\textbf{Component}} 
& \multicolumn{2}{c|}{\textbf{Gym99}} 
& \multicolumn{2}{c|}{\textbf{Gym288}} 
& \textbf{Diving48} \\
\cline{1-7}
Cross-attn. & Sparse MoE
& Top-1 & MCA 
& Top-1 & MCA 
& Top-1 \\
\hline

\xmark & \cmark
& 95.3 & 95.2 
& 92.4 & 73.6 
& 92.1 \\

\cmark & \xmark
& 95.8 & 95.2 
& 92.7 & 74.1 
& 92.1 \\

\hline

\cmark & \cmark
& \textbf{97.1} & \textbf{96.4} 
& \textbf{94.3} & \textbf{76.2} 
& \textbf{92.9} \\

\hline
\end{tabular}
\vspace{-1.5em}
\end{table}

\noindent\textbf{Effect of cross-attention and sparse experts.} Table~\ref{tab:ablation_component} isolates the two core components of FineX by comparing: (i) \emph{MoE only}, where cross-attention is removed, and stream representations are routed directly through the sparse expert layer; (ii) \emph{cross-attention only}, where the MoE is replaced by a shared dense transformation; and (iii) the full model. Both components help and have complementary effects. On Gym288, removing cross-attention reduces performance from 94.3/76.2 to 92.4/73.6 Top-1/MCA, while removing the sparse MoE gives 92.7/74.1. Cross-attention yields the larger individual gain, consistent with its role in exchanging evidence across RGB, pose-heatmap, and skeletal-graph streams. The sparse MoE then adds further improvement, indicating that conditional refinement remains useful after the streams are contextualised.

The same pattern appears on Gym99 and Diving48. On Gym99, both ablated variants reach 95.2 MCA, while the full model achieves 96.4, indicating that the modules are jointly beneficial even on a less long-tailed benchmark. On Diving48, both single-component variants obtain 92.1 Top-1, whereas FineX reaches 92.9. Thus, neither cross-attention nor sparse routing alone fully exploits the fine-grained kinematic structure of diving actions; their combination yields the most consistent gains across datasets.

\noindent\textbf{Cross-stream complementarity.}
Table~\ref{tab:ablation_stream} assesses whether the three evidence streams are redundant or complementary. Single-stream models are strong but dataset-dependent: pose heatmaps perform best on Gym99, RGB on Gym288 and Diving48, and skeletal graphs remain competitive but drop on the long-tailed Gym288. Thus, no single representation dominates across all fine-grained settings. Pairwise fusion clearly outperforms single streams, especially on Gym288: the best single stream reaches 89.1 Top-1 and 68.0 MCA, while the RGB+skeletal graph attains 93.6 Top-1 and 74.8 MCA. This gain shows that appearance and graph-based kinematics provide complementary cues: RGB supplies visual context and local motion, while the skeletal graph encodes inter-joint structure and temporal body dynamics. The RGB+skeletal graph is also the strongest pair on all three benchmarks, indicating that these two streams are the least redundant. 

The full three-stream model further improves over the best pair on every dataset. Gains are small on Gym99, where performance is near saturation, but larger on Gym288 MCA (+1.4) and Diving48 Top-1 (+1.2). This demonstrates that the pose-heatmap stream adds spatial evidence even when it is not the best standalone modality. Heatmaps preserve image-plane joint and limb locations that the skeletal graph abstracts away, and this signal becomes useful when cross-attention conditions it on RGB appearance and graph topology. Overall, the results support FineX’s key motivation: dense pose geometry and skeletal topology should be treated as distinct evidence sources rather than merged into a single pose modality.

\begin{table}[t]
\centering
\caption{\textbf{Cross-stream complementarity}
R $\rightarrow$ RGB, P $\rightarrow$ Pose-Heatmap, S $\rightarrow$ Skeletal Graph. We compare single-stream, pairwise, and full three-stream fusion. The strongest pair is shaded, while full model is in \textbf{Bold}.}
\label{tab:ablation_stream}
\scriptsize
\setlength{\tabcolsep}{4pt}

\begin{tabular}{|ccc|cc|cc|c|}
\hline
\multicolumn{3}{|c|}{\textbf{Stream}} 
& \multicolumn{2}{c|}{\textbf{Gym99}} 
& \multicolumn{2}{c|}{\textbf{Gym288}} 
& \textbf{Diving48} \\
\cline{1-8}
R & P & S 
& Top-1 & MCA 
& Top-1 & MCA 
& Top-1 \\
\hline

\cmark & \xmark & \xmark 
& 93.6 & 91.5 
& 89.1 & 68.0 
& 87.5 \\

\xmark & \xmark & \cmark 
& 94.1 & 92.1 
& 85.6 & 62.8 
& 86.2 \\

\xmark & \cmark & \xmark 
& 94.6 & 92.5 
& 86.3 & 67.7 
& 82.7 \\

\hline

\xmark & \cmark & \cmark 
& 95.6 & 94.0 
& 87.9 & 68.5 
& 87.4 \\

\cmark & \cmark & \xmark 
& 96.3 & 94.9 
& 93.2 & 74.3 
& 89.6 \\

\rowcolor{gray!15}
\cmark & \xmark & \cmark 
& 96.8 & 95.9 
& 93.6 & 74.8 
& 91.7 \\

\hline

\cmark & \cmark & \cmark 
& \textbf{97.1} & \textbf{96.4} 
& \textbf{94.3} & \textbf{76.2} 
& \textbf{92.9} \\

\hline
\end{tabular}
\vspace{-1.5em}
\end{table}

\noindent\textbf{Effect of load-balancing.} Table~\ref{tab:ablation_lambda} analyses the coefficient $\lambda_{\mathrm{lb}}$ in Eq.~\eqref{eq:final_objective}. Adding the load-balancing term improves all metrics over cross-entropy alone, with the largest gain on Gym288 MCA (75.2 to 76.2 at $\lambda_{\mathrm{lb}}=0.1$), while Top-1 changes only slightly. This matches MCA’s higher sensitivity to class-wise failures and Top-1’s focus on frequent classes, so balanced expert utilisation matters most in the long-tailed setting. Performance peaks at $\lambda_{\mathrm{lb}}=0.1$ across all benchmarks. Increasing $\lambda_{\mathrm{lb}}$ to 0.2 hurts performance (Gym288 MCA: 76.2 to 75.7), suggesting that regulariser must be strong enough to avoid expert under-utilisation but not so strong that it impedes specialisation from sparse routing. We, therefore, set $\lambda_{\mathrm{lb}}=0.1$ in all main experiments.

\begin{table}[t]
\centering
\begin{minipage}{0.48\textwidth}
\centering
\caption{Effect of the load-balance loss coefficient $\lambda_{\mathrm{lb}}$ in Eq.~\eqref{eq:final_objective}. ($\lambda_{\mathrm{lb}}=0\rightarrow $ CE)  }
\label{tab:ablation_lambda}
\scriptsize
\setlength{\tabcolsep}{3pt}
\begin{tabular}{|l|cc|cc|c|}
\hline
\multirow{2}{*}{\textbf{$\lambda_{\mathrm{lb}}$}} & \multicolumn{2}{c|}{\textbf{Gym99}} & \multicolumn{2}{c|}{\textbf{Gym288}} & \textbf{Diving48} \\
\cline{2-6}
 & Top-1 & MCA & Top-1 & MCA & Top-1 \\
\hline
0   & 96.9 & 96.0 & 94.1 & 75.2 & 92.3 \\
0.05         & 97.0 & 96.1 & 94.1 & 75.8 & 92.5 \\
\textbf{0.1} & \textbf{97.1} & \textbf{96.4} & \textbf{94.3} & \textbf{76.2} & \textbf{92.9} \\
0.2          & 97.0 & 96.1 & 94.2 & 75.7 & 92.6 \\
\hline
\end{tabular}
\end{minipage}
\hfill
\begin{minipage}{0.48\textwidth}
\centering
\caption{Effect of top-$k$ expert selection with $N{=}16$ experts. ($k{=}16\rightarrow $ Dense)}
\label{tab:ablation_topk}
\scriptsize
\setlength{\tabcolsep}{3pt}
\begin{tabular}{|l|cc|cc|c|}
\hline
\multirow{2}{*}{\textbf{$k$}} & \multicolumn{2}{c|}{\textbf{Gym99}} & \multicolumn{2}{c|}{\textbf{Gym288}} & \textbf{Diving48} \\
\cline{2-6}
 & Top-1 & MCA & Top-1 & MCA & Top-1 \\
\hline
1     & 96.2 & 95.2 & 93.3 & 73.7 & 90.2 \\
4     & 97.0 & 96.1 & 94.0 & 74.6 & 91.9 \\
\textbf{8} & \textbf{97.1} & \textbf{96.4} & \textbf{94.3} & \textbf{76.2} & \textbf{92.9} \\
16  & 97.1 & 96.2 & 94.1 & 75.1 & 92.2 \\
\hline
\end{tabular}
\end{minipage}
\vspace{-1.5em}
\end{table}

\noindent\textbf{Effect of top-$k$ expert routing.} Table~\ref{tab:ablation_topk} varies the number of active experts per stream while fixing the expert pool to $N=16$. Setting $k=1$ gives hard single-expert routing, and $k=16$ activates all experts, removing sparsity. Performance improves from $k=1$ to $k=8$ (Gym288 MCA 73.7$\rightarrow $76.2), indicating that single-expert routing is too restrictive for FHAR, where multiple fine-grained cues must be combined within a stream. Dense activation is also suboptimal: increasing $k$ from 8 to 16 reduces Gym288 MCA from 76.2 to 75.1 and Diving48 Top-1 from 92.9 to 92.2. With all experts active, the router cannot enforce sparse, specialised transformations, and each token is processed by the full expert bank, weakening the intended conditional computation. The best trade-off is at $k=8$, which keeps a soft mixture of experts with sparse, content-dependent routing. We adopt $k=8$ as the default. Further ablation results for varying numbers of experts are provided in supplementary material.

\vspace{-0.5em}

\subsection{Qualitative Analysis}
\label{sec:qualitative}


\begin{figure*}[t]
\centering
\setlength{\tabcolsep}{2pt}
\renewcommand{\arraystretch}{1.0}
\begin{tabular}{cccc}
\includegraphics[width=0.23\textwidth]{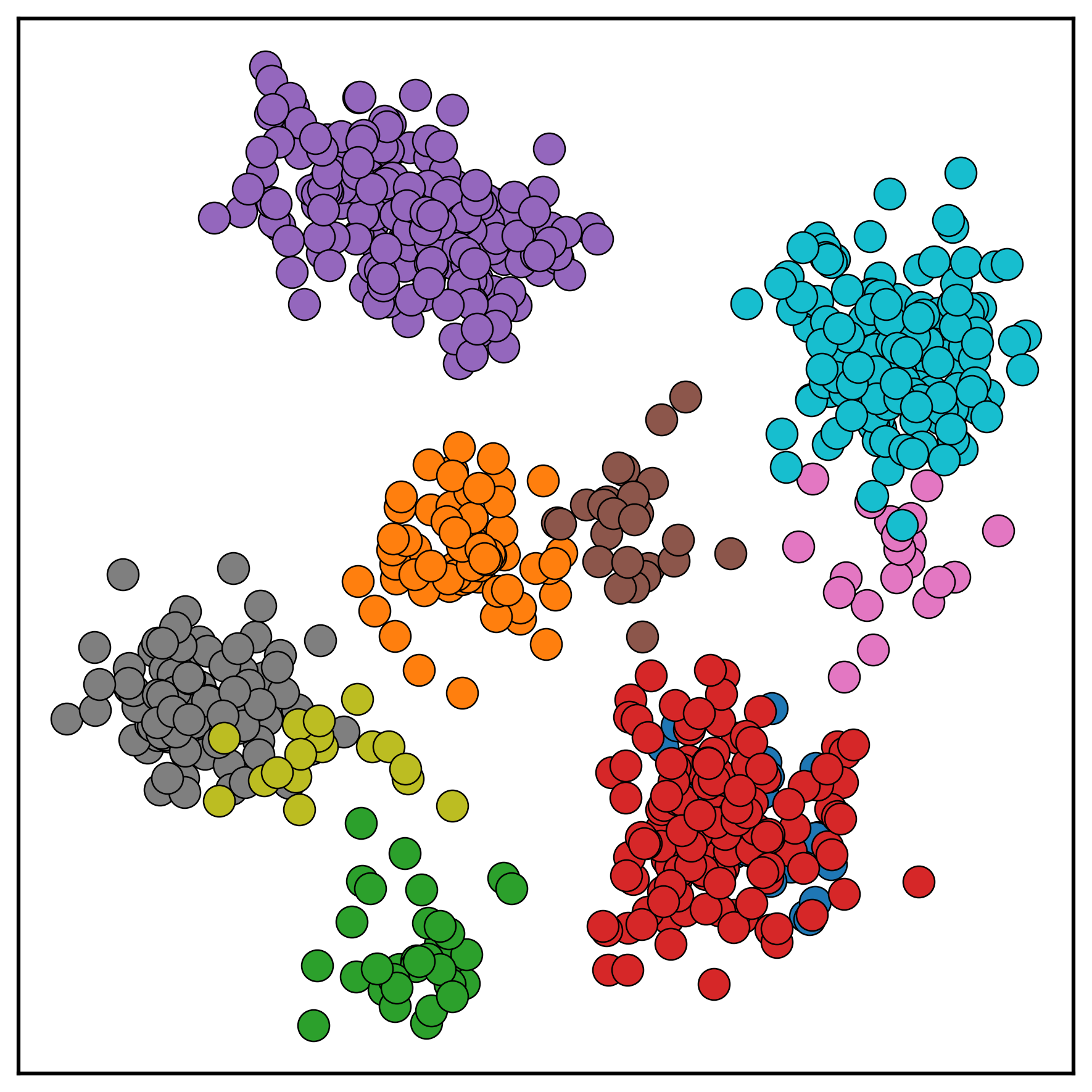} &
\includegraphics[width=0.23\textwidth]{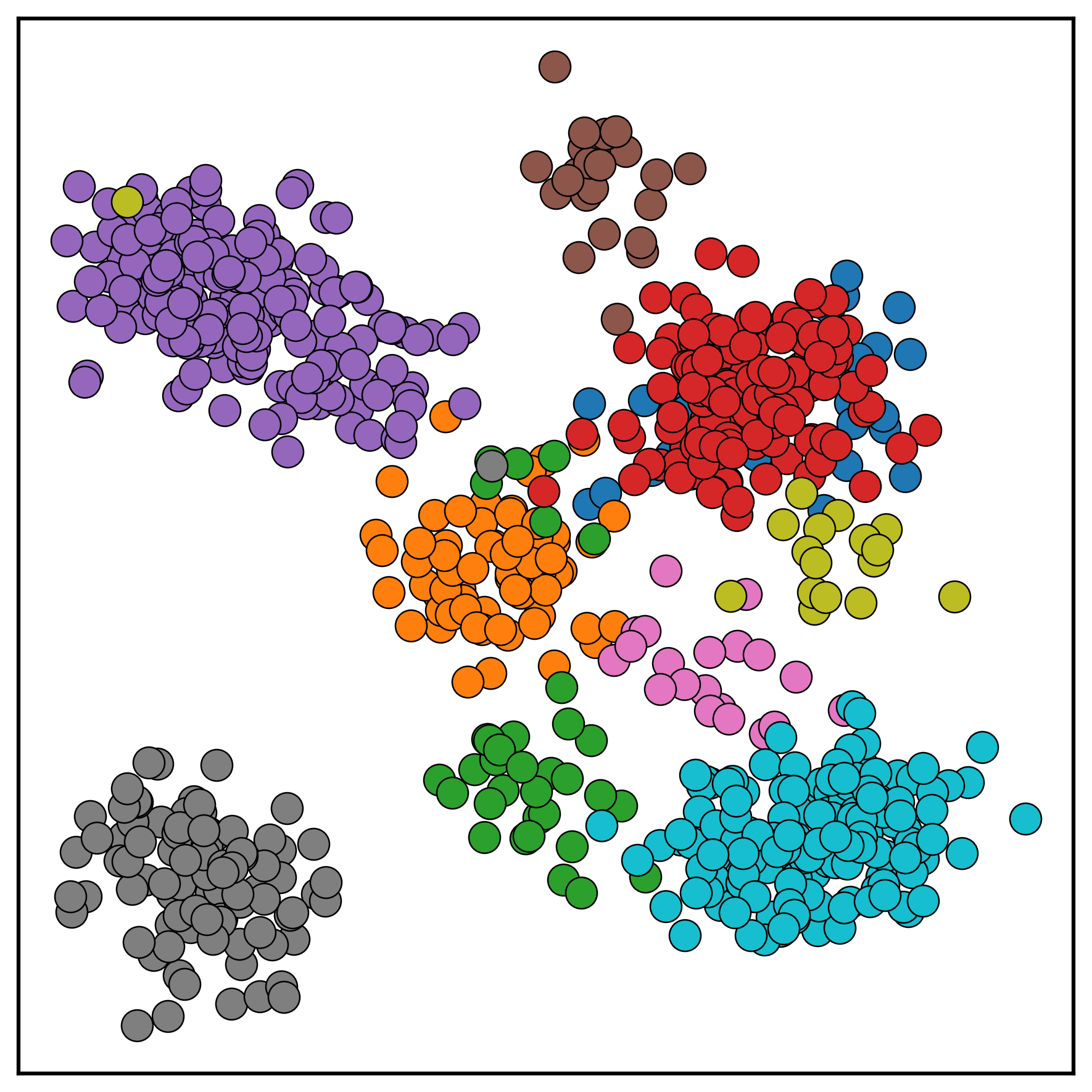} &
\includegraphics[width=0.23\textwidth]{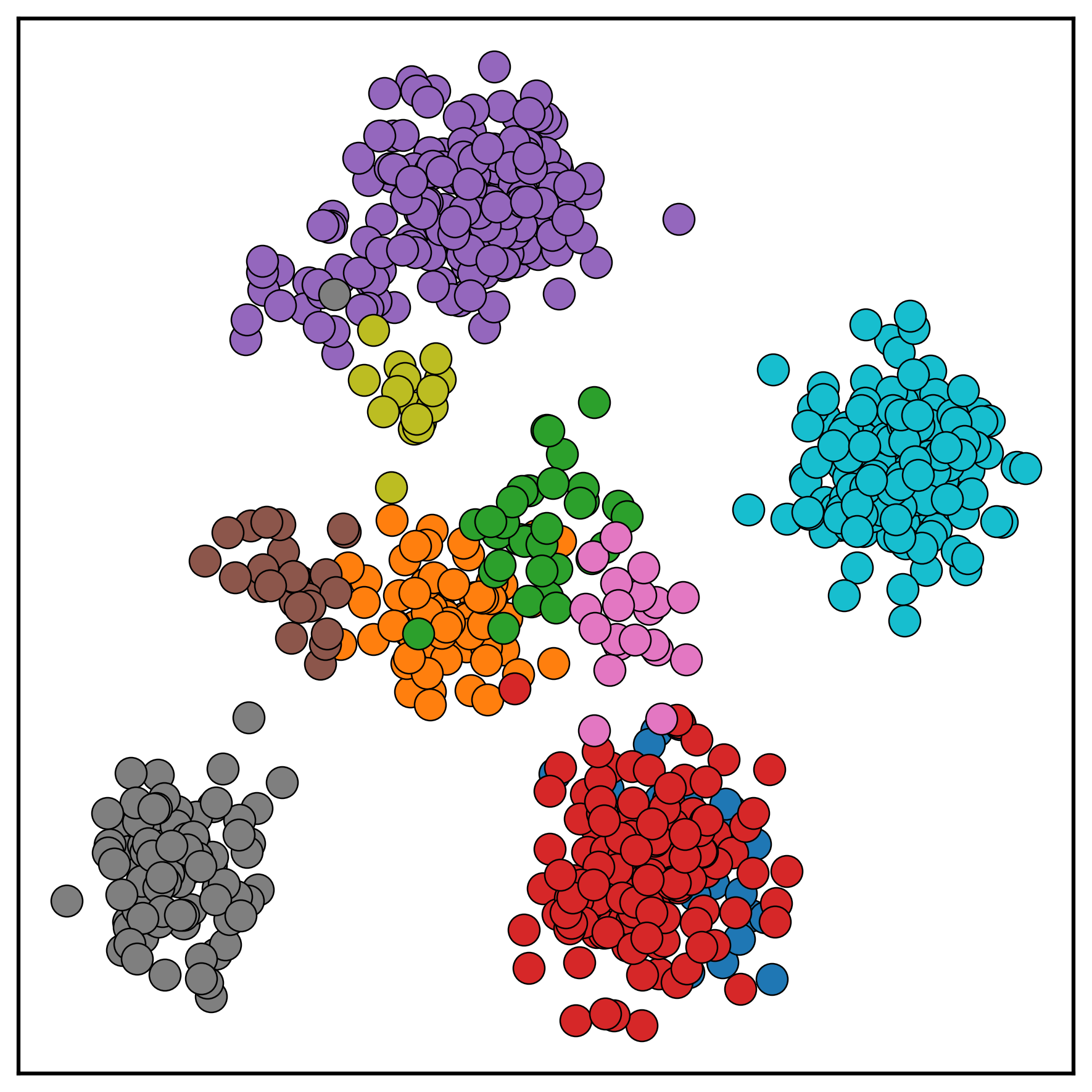} &
\includegraphics[width=0.23\textwidth]{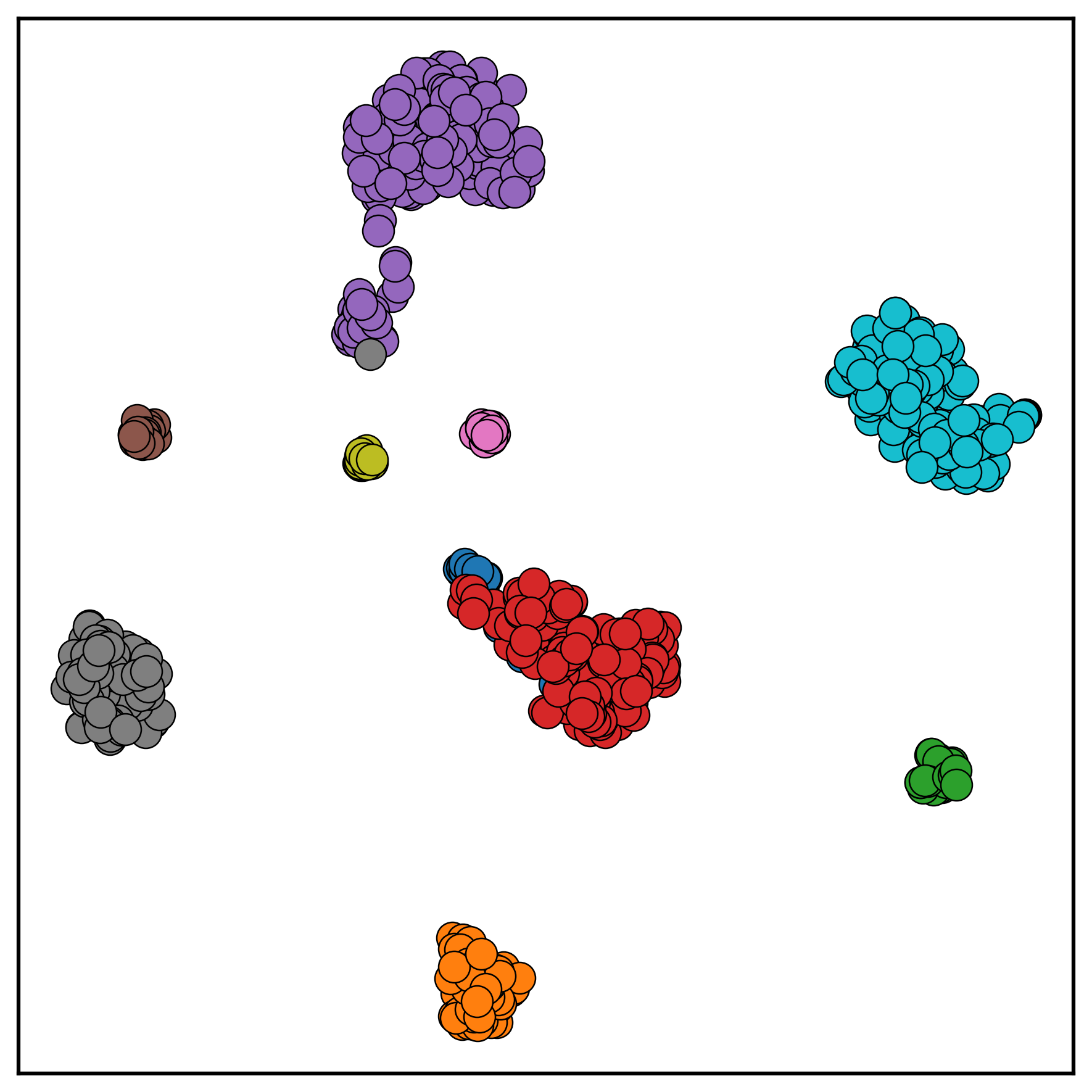} \\
{\small (a) RGB} & {\small (b) Pose heatmap} & {\small (c) Skeleton graph} & {\small (d) FineX} \\
\end{tabular}
\caption{\textbf{t-SNE visualisation of feature embeddings on Gym288.}
Embeddings from (a) RGB R(2+1)D, (b) pose-heatmap PoseC3D--SlowOnly,
(c) skeletal-graph ST-GCN++, and (d) the final FineX representation, for 10
randomly selected classes. Single-stream embeddings exhibit noticeable
class overlap, while FineX forms more compact and better separated
clusters.}
\label{fig:tsne-comprehensive}
\vspace{-1em}
\end{figure*}

\noindent\textbf{Embedding separability.} Fig.~\ref{fig:tsne-comprehensive} shows t-SNE embeddings \cite{vandermaaten2008tsne} for 10 randomly chosen classes for Gym288, illustrating the limits of single evidence streams. RGB features capture appearance and local motion but confuse classes with similar visual context and different execution details. Pose-heatmap features preserve image-plane joint locations but can merge classes with similar instantaneous poses. Skeletal-graph features model inter-joint topology and temporal kinematics but discard dense visual context useful for distinguishing visually similar phases. FineX yields tighter within-class clusters and clearer inter-class separation on all three datasets. This matches the roles of the two fusion stages: pairwise cross-attention reduces single-stream ambiguity by conditioning each representation on the other two streams, and streamwise sparse MoE provides content-adaptive nonlinear refinement. While t-SNE is only a qualitative projection and not a standalone metric, its geometry aligns with the quantitative results in Tables~\ref{tab:merged_comparison} and~\ref{tab:ablation_stream}, indicating that the fused representation is more discriminative than any individual stream. Furthermore, t-SNE visualisation for Gym99 and Diving48  is provided in supplementary section.


\begin{figure*}[t]
    \centering
    \setlength{\tabcolsep}{2pt}
    \renewcommand{\arraystretch}{1.0}
    \begin{tabular}{ccc}
        \includegraphics[width=0.33\textwidth]{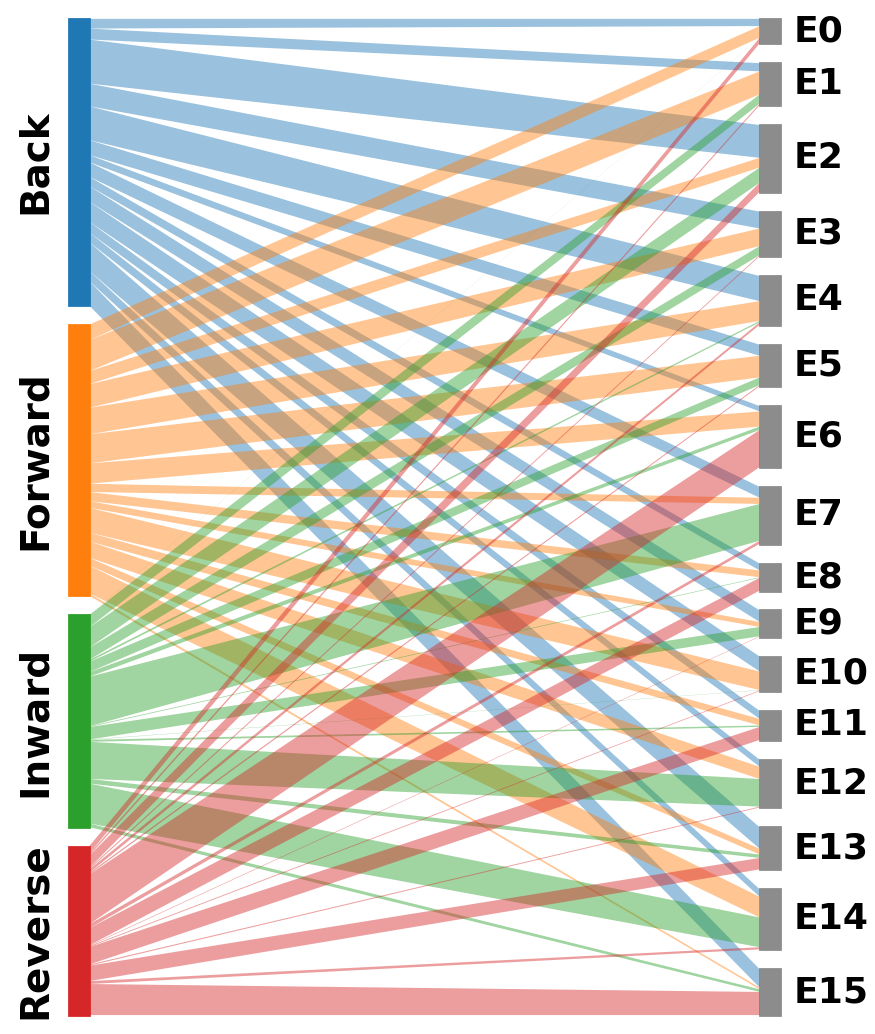} &
        \includegraphics[width=0.33\textwidth]{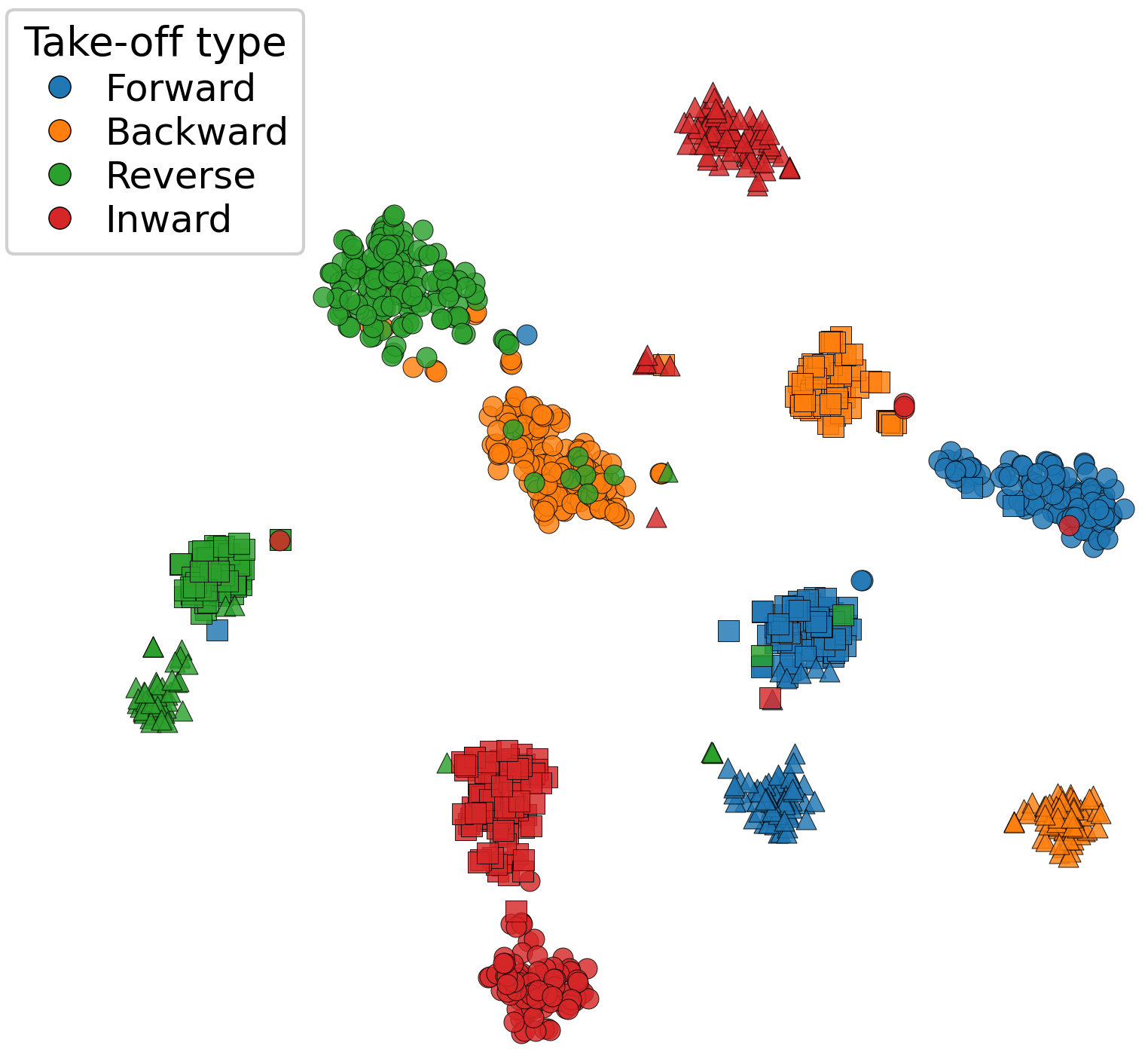} &
        \includegraphics[width=0.33\textwidth]{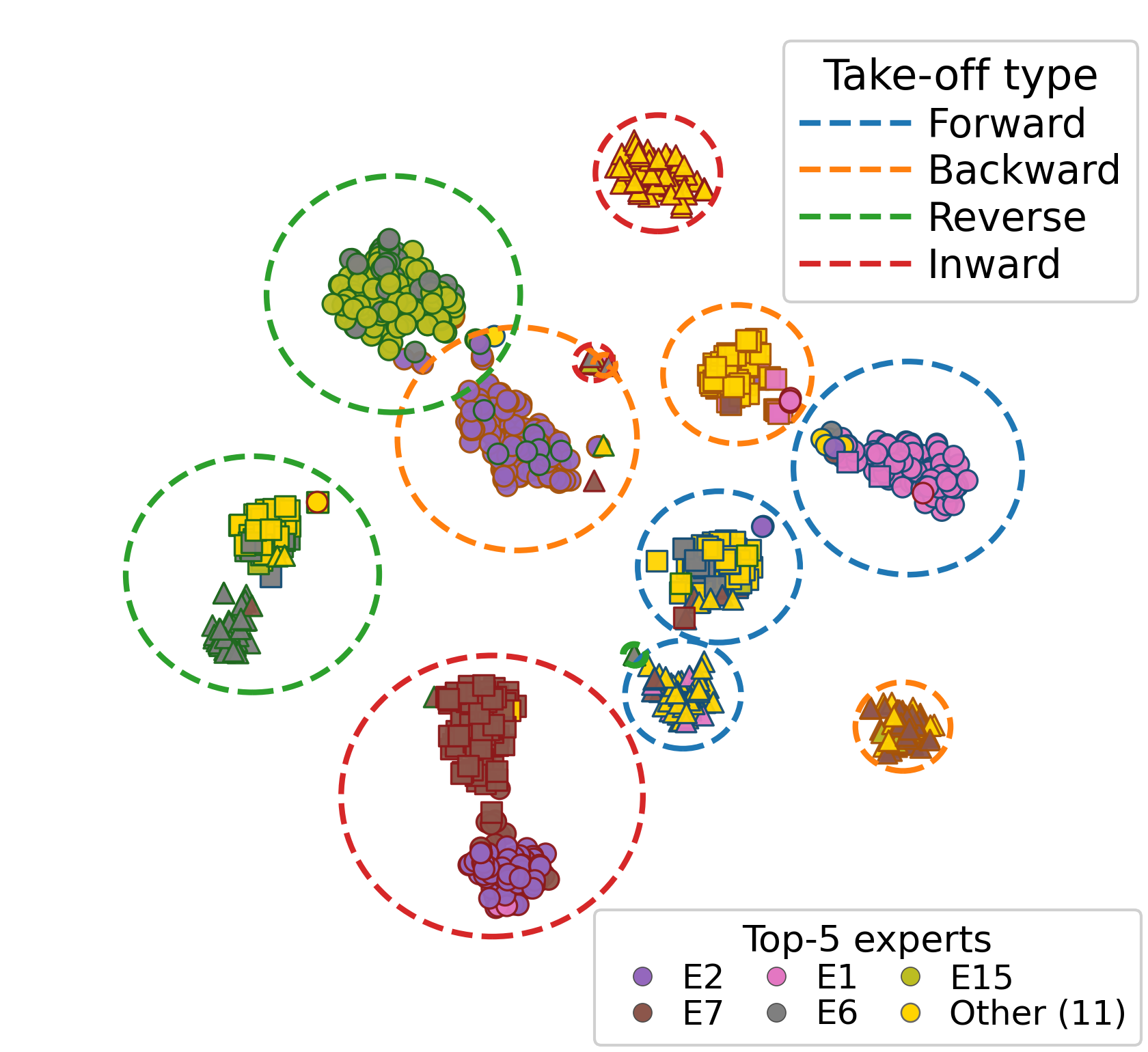} \\
        {\small (a)} & {\small (b)} & {\small (c)} \\
    \end{tabular}

\caption{\textbf{Expert routing on Diving48 at both coarse and fine-grained resolutions.} (a) Sankey diagram summarising dominant expert selections by take-off type, with flow width indicating the number of routed tokens. (b--c) t-SNE visualisation of concatenated stream-level routing weights, with three fine-grained classes per take-off type (up to 150 clips each); the embedding is identical in both panels, differing only in colour/fill. (b) Colour encodes take-off type, and marker shape encodes fine-grained class. (c) Fill indicates the dominant expert (the five most frequently selected experts shown individually, all others grouped as \emph{Others}); edge colour shows take-off type, and dashed outlines indicate the approximate cluster boundary for each take-off type.}
\label{fig:family_expert_routing}
\vspace{-.5em}
\end{figure*}

\noindent\textbf{Expert routing on Diving48.}
Fig.~\ref{fig:family_expert_routing} analyses expert routing at both coarse and fine-grained levels. Fig.~\ref{fig:family_expert_routing}(a) shows a Sankey diagram that aggregates dominant-expert assignments by take-off type, with flow width indicating the number of routed tokens. The routing pattern follows motion structure: board-facing \textbf{Reverse} and \textbf{Inward} take-offs rely on a smaller subset of dominant experts than \textbf{Forward} and \textbf{Backward}. Importantly, no expert is \emph{exclusive} to a single take-off type. Instead, experts are shared across categories, capturing reusable motion cues rather than simply memorising high-level labels.

Fig.~\ref{fig:family_expert_routing}(b--c) then focuses on three randomly chosen fine-grained classes per take-off type (each with at least 20 validation clips), summarising routing over the five most frequently used experts plus an \emph{Others} group. Even within this restricted expert pool, routing remains class-specific: distinct fine-grained classes under the same take-off type are typically dominated by different experts from the pool. Inward classes predominantly route to \textbf{E2} and \textbf{E7}, with a minor fraction going to \emph{Others}. Forward traffic is split mainly between \textbf{E1} and \emph{Others}, with a smaller contribution from \textbf{E6}. Reverse mainly routes to \textbf{E15}, then \textbf{E6}, with a small portion in \emph{Others}. Backward is dominated by \textbf{E2}, with a smaller share in \emph{Others} and a few clips in \textbf{E1}. Clips belonging to the same class form compact, largely well-separated clusters, with only occasional outliers beyond their class boundary. Altogether, the three panels indicate that take-off type restricts the subset of experts a clip uses, but does not reduce routing to a single shared expert; fine-grained classes within the same take-off type still preferentially route to different experts within that subset.

\begin{figure}[t]
\centering
\includegraphics[width=0.95\linewidth]{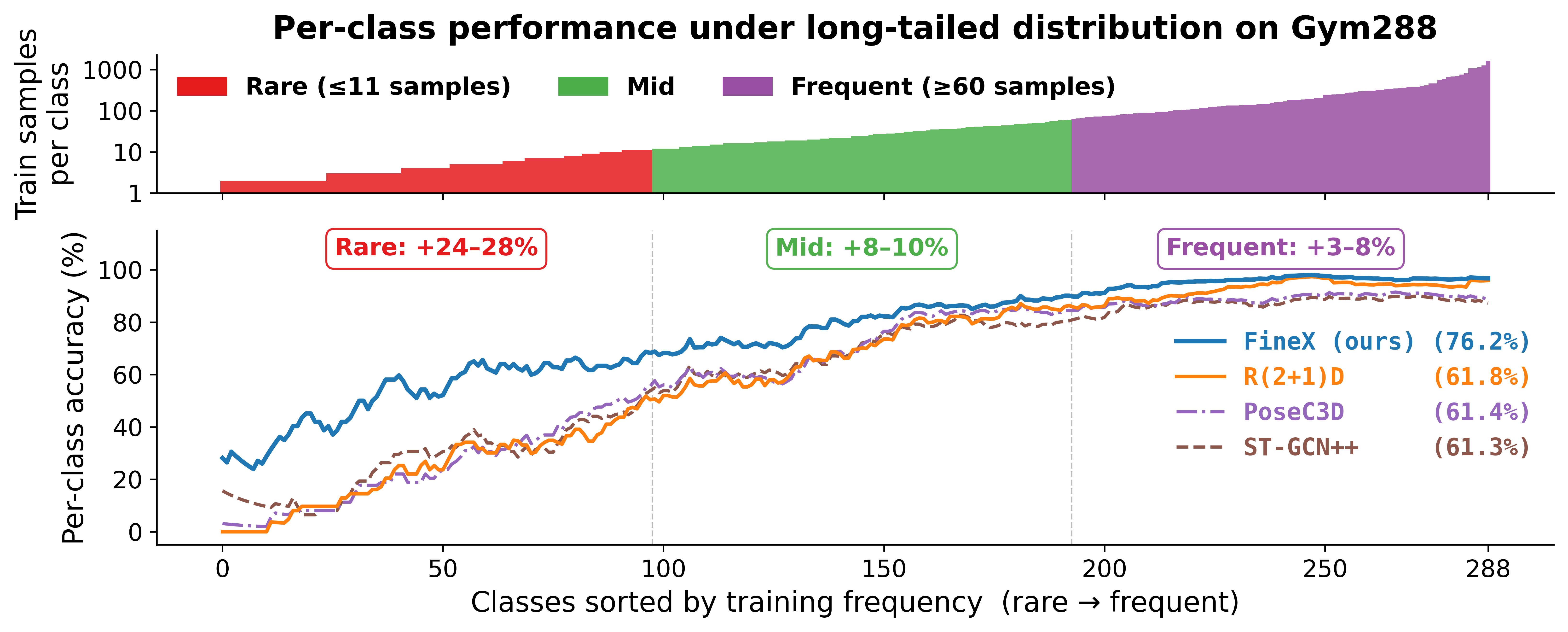}

\caption{\textbf{Per-class accuracy on Gym288 sorted by training
frequency.} Top: class-frequency histogram. Bottom: per-class accuracy of FineX vs. the three single-stream backbones. FineX gives the largest gains
on rare classes.}
\label{fig:longtail_gain}
\end{figure}




\noindent\textbf{Long-tail per-class behaviour.} Fig.~\ref{fig:longtail_gain} further analyses Gym288 by sorting classes from rarest to most frequent by number of training samples. The frequency histogram confirms the benchmark’s severe long-tailed distribution, with many categories containing at most 11 training instances. Thus MCA is a more \emph{informative} metric than Top-1 accuracy, as it measures class-wise recognition instead of being dominated by frequent categories.

The single-stream backbones reach similar MCA and saturate near 61\%: R(2+1)D at 61.8\%, SlowOnly at 61.4\%, and ST-GCN++ at 61.3\%. FineX boosts MCA to \textbf{76.2\%}, a \textbf{+14.4} gain over the best single-stream model. This improvement is strongly frequency-dependent: FineX achieves the largest gains on rare classes (\textbf{+24--28}), followed by mid-frequency classes (\textbf{+8--10}), with smaller but consistent gains on frequent classes (\textbf{+3--8}).

This pattern is key. For frequent classes, each stream sees enough data to learn discriminative cues, so fusion adds only modest gains. For rare classes, relying on a single representation is fragile: RGB may miss joint-level structure, pose heatmaps may miss topological relations, and skeletal graphs may miss visual context. FineX addresses by fusing appearance, dense pose geometry, and skeletal topology before sparse expert refinement. Thus, long-tail analysis reinforces the main 
finding on Gym288: FineX boosts not only average accuracy but also class-balanced performance for scarce training. The same complementarity yields robustness to noisy pose: PoseC3D's heatmaps better tolerate joint noise than raw coordinates~\cite{duan2022revisiting}, while dual-pose decomposition adds strength over individual stream. Crucially, cross-attention weighting attenuates corrupted pose queries, suppressing unreliable streams rather than propagating their errors.



\vspace{-.5em}

\subsection{Computational Efficiency}
\label{sec:efficiency}
FineX separates backbone representation learning from cross-modal fusion. In Stage~1, the three modality-specific backbones are fine-tuned individually and then frozen; R(2+1)D-34, PoseC3D with SlowOnly-R50 Backbone, and ST-GCN++ together have 67.7M parameters. In Stage~2, we train only the 7.5M-parameter fusion module on pre-cached features. Because it operates on compact feature vectors ($\mathbf{f}_r,\mathbf{f}_s \in \mathbb{R}^{512}$ and $\mathbf{f}_g \in \mathbb{R}^{256}$), its cost is independent of the backbones’ spatial and temporal resolution, making Stage~2 training lightweight and avoiding repeated backpropagation through the video backbones.

\noindent\textbf{Comparison with parameter-efficient video adaptation.}
Table~\ref{tab:efficiency} compares FineX with AIM~\cite{yang2024aim} on Diving48 in terms of total parameters, Stage~2 tunable parameters, single-clip GFLOPs and Top-1 accuracy. As in previous adapter-based work, the \emph{Tunable} column counts only parameters updated in Stage~2. FineX has 75.2M total parameters (three frozen backbones plus fusion), fewer than AIM ViT-B/16 (97M) and far fewer than AIM ViT-L/14 (341M). For inference, FineX uses 682 GFLOPs per clip, versus 809 GFLOPs for AIM ViT-B/16 and 3,736 GFLOPs for AIM ViT-L/14.


Despite invoking three backbones, FineX incurs only 682 GFLOPs per clip, which is $5.5\times$ less than AIM ViT-L/14 (3{,}736) and below AIM ViT-B/16, and since three streams share no data dependency, they can execute in parallel on separate devices, without adding wall-clock latency. It also attains higher Diving48 accuracy than both AIM variants with fewer tunable parameters and lower GFLOPs, showing that improvements come from cross-modal decomposition and efficient fusion of experts rather than simply scaling model capacity.

\begin{table}[t]
\centering

\caption{Diving48 Efficiency comparison. Tunable parameters refer Stage2 fusion parameters. GFLOPs are measured for single-clip inference. R$\rightarrow$RGB, P$\rightarrow$Pose-Heatmap.
}

\label{tab:efficiency}
\small
\setlength{\tabcolsep}{3.5pt}
\begin{tabular}{|l|c|c|c|c|c|}
\hline
\textbf{Method} & \textbf{Total (M)} & \textbf{Tunable (M)} & \textbf{GFLOPs} & \textbf{Top-1} & \textbf{Input} \\
\hline
AIM ViT-B/16~\cite{yang2024aim} & 97 & 11 & 809 & 88.9 & R \\
AIM ViT-L/14~\cite{yang2024aim} & 341 & 38 & 3{,}736 & 90.6 & R \\
\hline
\textbf{FineX (ours)} & \textbf{75.2} & \textbf{7.5} & \textbf{682} & \textbf{92.9} & R+P \\
\hline
\end{tabular}
\par\smallskip
\vspace{-1em}
\end{table}

\vspace{-0.5em}

\section{Conclusion}
We introduced \textbf{FineX}, a cross-attentive sparse expert framework for fine-grained human action recognition. It models fine-grained evidence as three complementary streams: RGB appearance, pose-heatmap geometry, and skeletal-graph topology, while preserving stream identity. Instead of simple concatenation or late fusion, FineX preserves stream identity, enables pairwise cross-stream reasoning, and streamwise latent sparse expert routing for content-adaptive refinement. Cross-attention chooses what evidence to exchange between streams, and sparse experts determine how each resulting representation is transformed.

Experiments on Gym99, Gym288, and Diving48 show that this structured fusion is consistently effective. On the long-tailed Gym288 benchmark, FineX achieves \textbf{94.3\%} Top-1 accuracy and \textbf{76.2\%} MCA, improving the previous best MCA by \textbf{+7.6} and the strongest vision-language baseline by \textbf{+12.2}, with the largest gains on rare and mid-frequency classes. This suggests that complementary visual, pose, and graph cues offset the weaknesses of any single stream under data scarcity. On Gym99 and Diving48, FineX also reaches state-of-the-art performance, indicating that the design benefits both balanced fine-grained discrimination and motion-structured recognition.

Importantly, FineX obtains these gains without textual supervision or large-scale vision-language pre-training, indicating that, for FHAR, explicitly modeling visual and pose structure can rival scaling semantic supervision. However, FineX has limitations: it needs three backbone forward passes at inference. Future work could reduce computation via shared or lightweight pose representations, improve robustness to noisy keypoints, and test whether language priors add complementary gains for visually ambiguous classes.

\section*{Acknowledgments}
This work was supported by UK Research and Innovation (UKRI) - Engineering and Physical Sciences Research Council (EPSRC) under Grant EP/X028631/1 (ATRACT Project). 

\clearpage  


%
%
\bibliographystyle{splncs04}
\bibliography{main}
\end{document}